\documentclass[runningheads]{llncs}
\usepackage{graphicx}
\usepackage{multirow} 
\usepackage{fontawesome}

\begin{document}
%
\title{Synthesizing Realistic Data for Table Recognition} 
%
%
\author{Qiyu Hou\inst{1}\orcidID{0009-0009-4150-9907} \and
Jun Wang\inst{1}\orcidID{0000-0002-9515-076X}(\faEnvelopeO) \and
Meixuan Qiao\inst{2} \and
Lujun Tian\inst{1} 
}
\authorrunning{Q. Hou et al.}
%
\institute{iWudao Tech\\ 
\email{\{houqy,jwang,tianlj\}@iwudao.tech} \\
\and
Huazhong University of Science and Technology \\
\email{qiaomeixuan@hust.edu.cn}
}

\maketitle              
\begin{abstract}
To overcome the limitations and challenges of current automatic table data annotation methods and random table data synthesis approaches, we propose a novel method for synthesizing annotation data specifically designed for table recognition. This method utilizes the structure and content of existing complex tables, facilitating the efficient creation of tables that closely replicate the authentic styles found in the target domain.
By leveraging the actual structure and content of tables from Chinese financial announcements, we have developed the first extensive table annotation dataset in this domain. We used this dataset to train several recent deep learning-based end-to-end table recognition models. Additionally, we have established the inaugural benchmark for real-world complex tables in the Chinese financial announcement domain, using it to assess the performance of models trained on our synthetic data, thereby effectively validating our method's practicality and effectiveness.
Furthermore, we applied our synthesis method to augment the FinTabNet dataset, extracted from English financial announcements, by increasing the proportion of tables with multiple spanning cells to introduce greater complexity. Our experiments show that models trained on this augmented dataset achieve comprehensive improvements in performance, especially in the recognition of tables with multiple spanning cells.

\end{abstract}
\section{Introduction}

Tables, serving as a vital carrier of data, are prevalent across a wide range of digital documents. They efficiently store and display data in a compact and lucid format, encapsulating an immense volume of valuable information. 
%
However, recognizing the structures of tables within digital documents, such as PDF and images, and subsequently extracting structured data, present significant challenges due to the complexity and diversity of their structure and style~\cite{ComplexTable-ACL-2023}.
In recent years, with the advancement of deep learning, new methodologies have surfaced, leading to significant progress in table recognition~\cite{SPLERGE-2019,TabStruct-Net-2020,LGPMA-2021,tablemaster-2021,TableFormer-CVPR-2022}. Deep learning-based methods are adept at managing complex table structures and diverse styles more effectively. However, their reliance on large-scale, high-quality annotated table datasets for model training is pronounced~\cite{TableFormer-CVPR-2022,ComplexTable-ACL-2023}. The scarcity of comprehensive and intricately detailed, publicly accessible datasets emerges as a substantial barrier, impeding the further advancement of table structure recognition. 
To create large-scale datasets for table recognition, some researchers have initially started by utilizing specific repositories of scientific papers or financial reports. Each document in these repositories contains tables that correspond to some form of structured source codes (such as LaTeX, XML, HTML). They facilitate large-scale annotation of table recognition data by automatically mapping the displayed tables to their corresponding structured source codes.
The existing large-scale, real-world table annotation datasets~\cite{PubTables-1M-CVPR-2022,PubTabNet-ECCV-2020,FinTabNet-WACV-2021,TableBank-LREC-2020,Table2Latex-ICDAR-2019,TabLeX-ICDAR-2021} are few but have all been constructed using similar methodologies. However, the applicability of these methods for creating table annotation data is notably limited, as only a select number of fields have access to document repositories where structured source codes correspond to rendered tables. Given the substantial variations in table structures and styles across different domains and languages, the table styles featured in these datasets tend to exhibit similarities. This similarity poses challenges when attempting to apply these datasets to a broader range of domains~\cite{TableFormer-CVPR-2022,ComplexTable-ACL-2023}.
These automatically annotated datasets frequently contain a significant number of annotation errors. For instance, in our sampling of over 10,000 tables from the FinTabNet dataset, we found that approximately 9\% had obvious annotation errors. 
Furthermore, the annotation information required varies among different table recognition methods, and some of these table datasets do not completely fulfill the annotation requirements of several prevalent table recognition models~\cite{PubTables-1M-CVPR-2022}. 
To adapt more efficiently to a wider range of application domains, some researchers are exploring methods for synthesizing annotated data for table recognition. These methods largely depend on predefined structural templates and randomly selected text to generate table structures and fill in content. Moreover, they employ a web browser engine to render the synthesized tables and add annotations, following predefined style templates. However, tables rendered with HTML and CSS offer limited visual fidelity and often fall short of replicating the complex or unique appearances of tables in documents like PDFs. Consequently, they lack the richness and complexity required to accurately simulate the intricate table structures encountered in real-world scenarios.
%

\begin{figure}
\includegraphics[width=\textwidth]{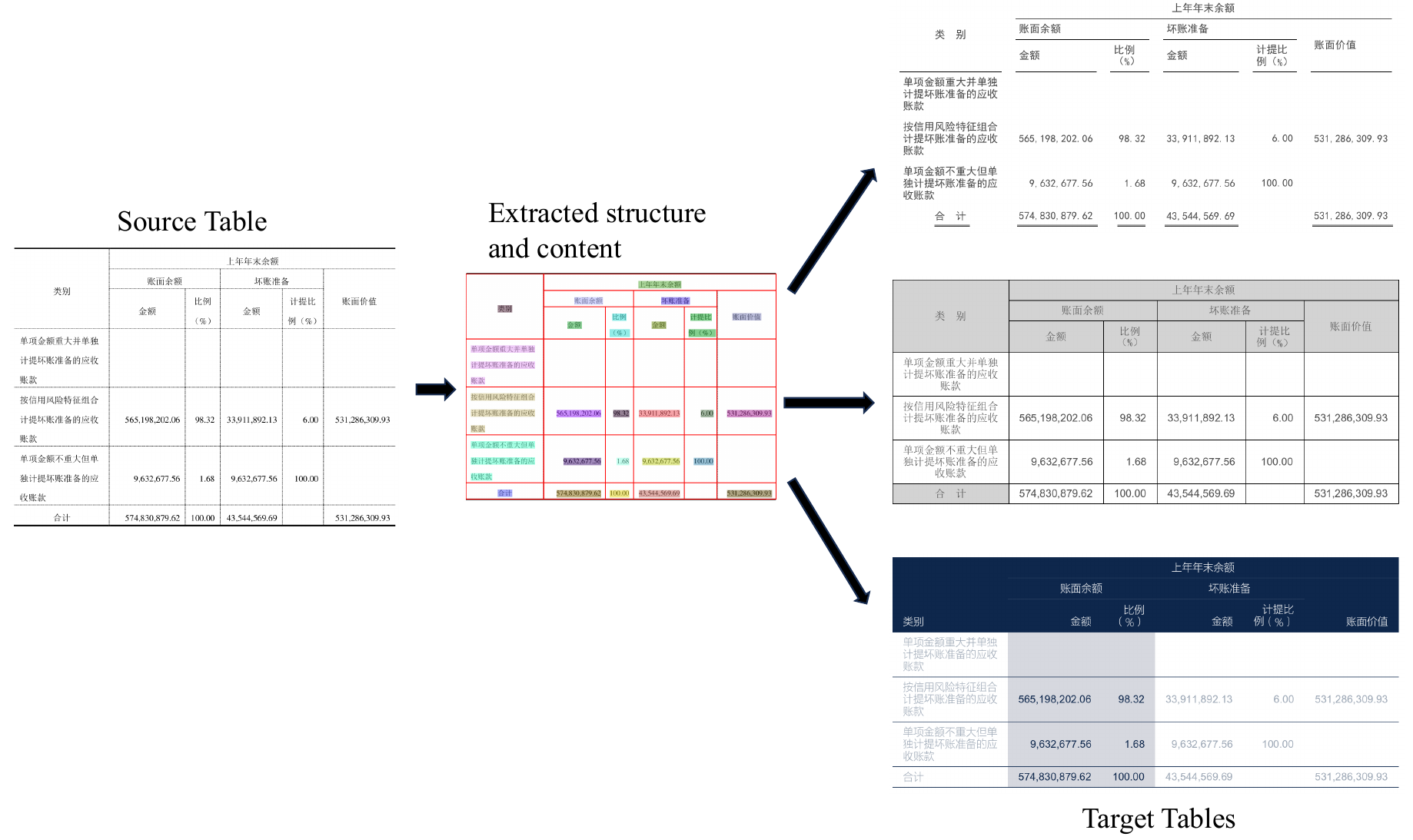}
\caption{The transformation of an original bordered table into two borderless tables and one bordered table with different style} \label{figure-table-style-transfer}
\end{figure}

%
%
After careful analysis of real-world tables, we discovered that generating appropriate templates for tables with complex structures, such as those found in the financial sector, is not easily achieved through random methods. The structures produced randomly often differ significantly from real complex tables, and the text randomly inserted into these tables usually deviates greatly from the context and semantics of actual tables.
We have also observed that in specific application domains, such as finance, the primary challenge in table recognition stems from the fact that many tables, despite having similar structures and content, display significantly varied presentation styles. To address this issue, we propose a method that utilizes the structure and content of existing complex tables to generate high-quality, realistic synthetic datasets tailored to the target application domain.
%
%
%
%
Firstly, in many scenarios, we can relatively easily access the structure and content of numerous existing tables. Taking the financial domain as an example, tables in U.S. financial disclosures are available in HTML format, allowing direct access to the table structure and the text content within the cells. Although Chinese financial announcements in PDF format do not have corresponding HTML files, they contain a large number of bordered tables. Using PDF parsing tools to extract and analyze the lines within these bordered tables, it is relatively easy to deduce the table structure and the text content in each cell. 
Secondly, reflecting the actual distribution of tables in the target application domain, tables can be categorized and summarized. This involves extracting and documenting the various attributes associated with the presentation styles of each table within every category. These attributes are then stored in a profile and incorporated into the candidate style set for that category.
For a source table whose structure and content have been obtained, its corresponding table category can first be identified based on its content. From this category's candidate set, a profile is selected, and a small degree of randomness is introduced to the attributes of the selected profile to set the profile for the target table. This results in the transformation into a new table that is completely different in style from the original, yet still possesses a very realistic appearance.
For example, Fig.~\ref{figure-table-style-transfer} shows the transformation of an original bordered table into two borderless tables and one bordered table with different style, all of which have the same structure and content. Additionally, our method, based on image rendering techniques for synthesizing tables, is not constrained by the limitations of browser rendering. It can draw or paste real table appearance components from various documents, enabling the creation of more realistic complex table styles.

This paper makes the following contributions:

$\bullet$ A novel method for synthesizing annotation data specifically designed for table recognition has been proposed, which utilizes the structure and content of existing complex tables. This approach enables the straightforward synthesis of tables that closely resemble the actual table styles prevalent in the target domain, accompanied by comprehensive and complete annotation data.

$\bullet$ Utilizing the actual structure and content of tables from Chinese financial announcements, we synthesized the first large-scale table annotation dataset in the domain of Chinese financial announcements. On this basis, we trained several recent deep learning-based end-to-end table recognition models. Furthermore, we created the first benchmark for real-world complex tables in the Chinese financial announcement domain, which was used to evaluate the performance of models trained on synthetic data, thereby validating the practicality and effectiveness of the method proposed in this paper.

$\bullet$Additionally, the method proposed in this paper was used to augment the table dataset extracted from English financial announcements, known as FinTabNet, by synthesizing and increasing the proportion of more complex tables with multiple spanning cells. Experiments demonstrate that table recognition models trained on this enhanced dataset exhibit comprehensive improvements in performance, particularly in recognizing more complex tables that contain multiple spanning cells.

\section{Related Work}
\subsection{Large-Scale Table Recognition Datasets}
The current large-scale table recognition datasets are automatically annotated by utilizing a few document repositories, where the tables displayed in the documents can be associated with corresponding structured source codes.

TABLE2LATEX-450K~\cite{Table2Latex-ICDAR-2019}, TableBank~\cite{TableBank-LREC-2020}, and TabLeX~\cite{TabLeX-ICDAR-2021} are datasets derived from articles in the Arxiv repository, where tables in PDF documents correspond to LaTeX source codes. Additionally, TableBank~\cite{TableBank-LREC-2020} has crawled some Word documents from the internet, linking tables in these documents to Office XML codes. PubTabNet~\cite{PubTabNet-ECCV-2020} and PubTables-1M~\cite{PubTables-1M-CVPR-2022} were sourced from scientific papers in the PubMed Central Open Access (PMCOA) database, with tables in the PDF files paired with corresponding XML codes. Meanwhile, FinTabNet~\cite{FinTabNet-WACV-2021} compiled annual reports from S\&P 500 companies, featuring tables in the PDF documents that can be associated with corresponding HTML codes.
The datasets mentioned above are confined to scientific papers or financial reports, and models trained on them may not perform well in other domains.
As mentioned before, these automatically annotated datasets often contain a considerable number of annotation errors. These issues encompass a variety of problems, including inaccuracies in localizing table regions, omissions of table content in annotations, structural errors in annotating headers or table content, and inconsistencies in annotations for identical structures or content within tables, among others.
Except for PubTables-1M, the annotations in the other datasets do not fully cover the annotation requirements needed by several common table recognition models~\cite{PubTables-1M-CVPR-2022}. Specifically, both PubTabNet and FinTabNet lack annotations for the coordinates of cell bounding boxes~\cite{PubTabNet-ECCV-2020,FinTabNet-WACV-2021}. Additionally, TableBank and TabLeX provide only the overarching structure of the table, without annotations for the content and coordinates of text blocks within each cell~\cite{TableBank-LREC-2020,TabLeX-ICDAR-2021}.

\subsection{Randomly Synthesized Data for Table Recognition}
Another series of efforts to address the scarcity of large-scale table recognition datasets involves constructing table annotation datasets through the synthesis of HTML tables.

Qasim et al.~\cite{Qasim-ICDAR-2019} initially employed four types of table templates to create a synthetic table dataset based on HTML. TableGeneration~\cite{TableGeneration} further expanded the method, maintaining support for four table templates while introducing a broader range of configurable parameters, which include cell type, the number of rows and columns in a table, the quantity of merged cells, and the provision for colored cells.
WikiTableSet~\cite{WikiTableSet-ICPRAM-2023} constructed a Wikipedia table extractor to harvest tables (in HTML code format) from the Wikipedia dump, and then normalized these HTML tables to align with the PubTabNet format, which involved separating table headers from data and stripping CSS and style tags. 
SynthTabNet~\cite{TableFormer-CVPR-2022} took a parameter-driven approach to initially generate the table structure, detailing the total number of rows and columns, header row count, types of spans (including header-only, row-only, column-only, and both row and column spans), maximum span size, and the proportion of table area covered by spans. Appropriate content templates are then selected and combined with purely random text to produce synthetic content. A collection of styling templates is manually curated, and a style is randomly selected to determine the appearance of the synthesized table. 
In a similar vein, the ComplexTable dataset~\cite{ComplexTable-ACL-2023} is synthetically generated using an auto HTML table creator, which produces table images alongside corresponding structured HTML code. Notably, ComplexTable features a significantly higher frequency of complex tables compared to SynthTabNet, and it showcases a more diverse range of table styles within the dataset.
The aforementioned methods employ HTML for table synthesis and CSS to define table styles, followed by the use of a Web Browser engine to render the table images. However, this approach has its limitations. Often, it falls short in replicating the complex or distinctive visual effects that tables in documents, such as PDFs, typically present.

\subsection{Table Augmentation}
Data augmentation is also a common method for acquiring table data. TabAug~\cite{TabAug-ICDAR-2021} introduced a novel data augmentation technique that primarily relies on two fundamental operations: Replication and Deletion, applied to rows and columns. Umer et al.~\cite{PyramidTabNet-ICDAR-2023} developed augmentation techniques that encompass clustering, fusion, and patching of table images. Ichikawa~\cite{Ichikawa-ICDAR-2021} introduced novel label-invariant table augmentation techniques focused on the edge-based region, demonstrating their substantial impact, particularly when training with limited datasets. Liu et al.~\cite{Liu-CVPR-2022} enhanced existing datasets by employing two types of image distortion algorithms, aiming to simulate distractors introduced by the capture device. 

This paper is mainly concerned with the acquisition of large-scale table annotation data encompassing diverse styles. The data augmentation methods mentioned do not significantly alter table styles, hence they differ somewhat from the core issue addressed in this paper.

\section{Our Approach for Synthesizing Annotated Tables}

\begin{figure}
\includegraphics[width=\textwidth]{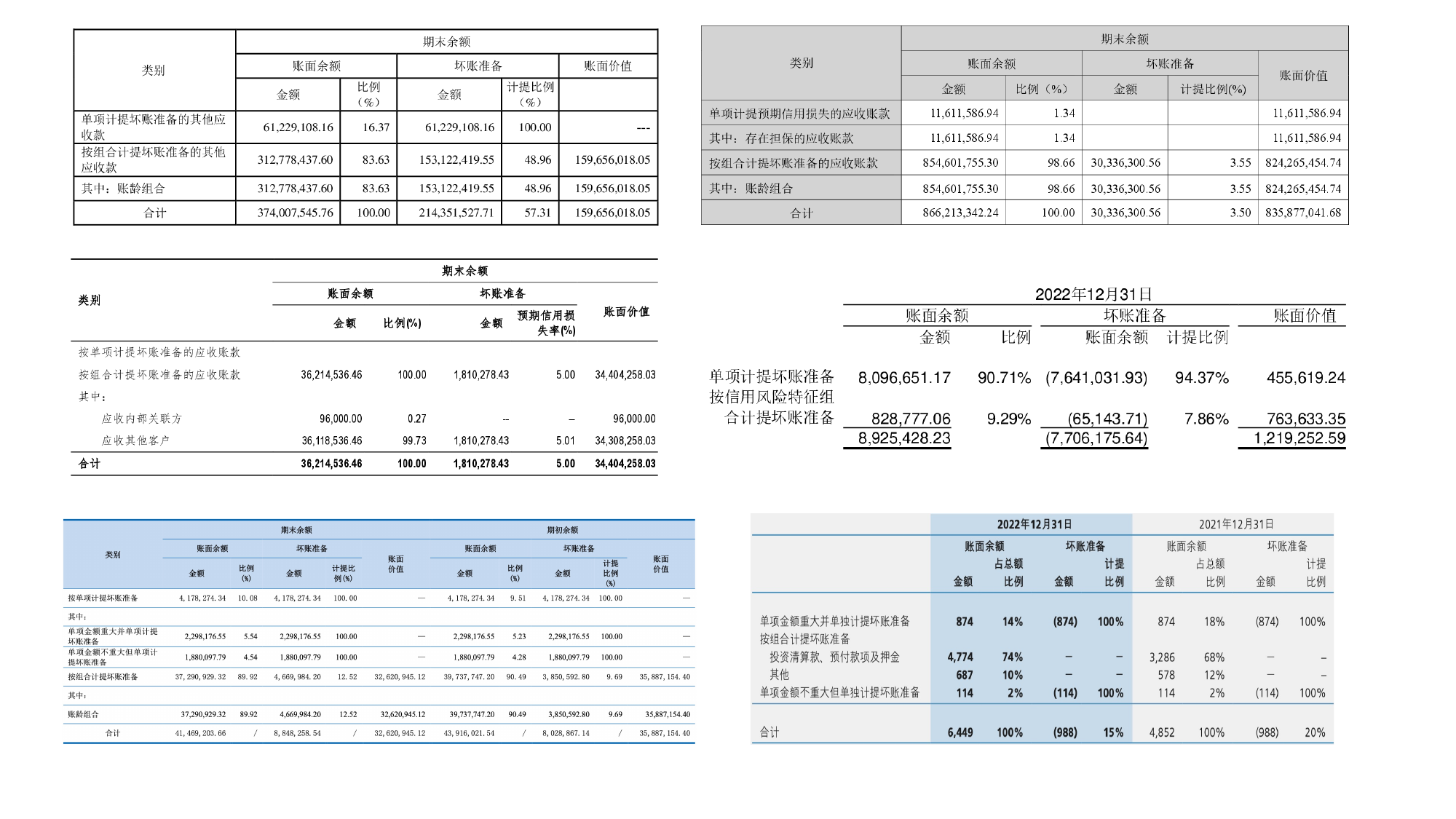}
\caption{Real examples of tables that have similar content but display vastly different visual styles.} \label{figure-content-style}
\end{figure}

If we aim to transform tables from documents into structured data for storage in databases or knowledge graphs within a specific domain, enabling further in-depth analysis and applications, it becomes essential to parse the structure of these tables to grasp their semantics. In many application domains, although the content of some tables across different documents may be very similar, the styles of these tables with similar content often exhibit significant differences.
Fig.~\ref{figure-content-style} shows six real examples of tables from the Financial Statements, specifically the ``Accounts Receivable - Disclosed by Provision Method" category. These tables, belonging to the same category, have similar content but vastly different visual styles.
Inspired by this observation, we propose synthesizing new target tables by leveraging the structure and content of existing complex tables, while applying completely different yet realistically plausible styles to these target tables. This approach ensures that the synthesized complex tables more closely resemble real-world scenarios than those produced by methods relying on randomly generated structures and content.

\begin{figure}
\includegraphics[width=\textwidth]{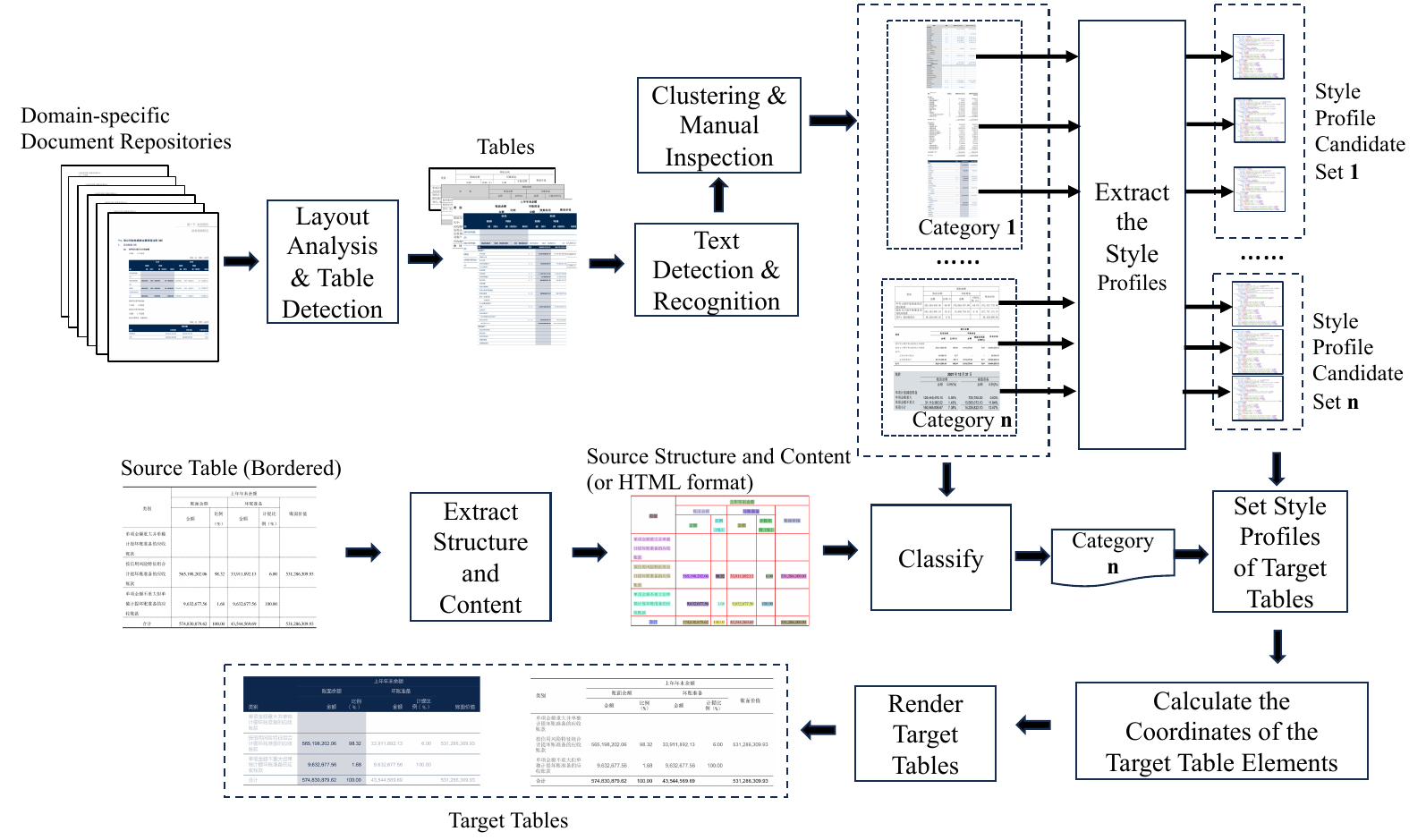}
\caption{System Structure of the table annotation data synthesis method} \label{Fig-System-structure}
\end{figure}

Fig.~\ref{Fig-System-structure} illustrates the overall structure of the table annotation data synthesis method proposed in this paper, which involves extensive handling of text and lines within tables. For tables generated via PDF coding, PDF parsing tools, such as pdfplumber,  offer a direct and efficient means to extract texts and lines. In contrast, for scanned tables, OCR can be used to detect and recognize text, and conventional methods such as the Hough Transform~\cite{Hough-transformation-1972,Hough-Transform-2000} or LSD~\cite{ipol.2012.gjmr-lsd}, as well as deep learning-based line detection techniques~\cite{Huang_2018_CVPR,ICCV-2019-zhou,zhang2019ppgnet,ECCV-2000-huang,AAAI-2022-Gu,CVPR-2021-Xu}, can be employed to identify lines within the tables.

\subsection{Analysis of the Real Table Distribution in the Target Domain}

To synthesize table annotation data that closely match the target application domain, a detailed investigation and analysis of the real table data's distribution within that domain are essential. In the document repository collected from the target domain, layout analysis and table detection are utilized to detect all included tables. OCR or PDF parsing tools are then used to detect and recognize the text content within these tables. Following this, tables are clustered based on their text content, grouping together those with similar content. Subsequently, manual inspection and adjustment are performed. This approach significantly facilitates the convenient and efficient identification of various critical table types within the application domain.
Through clustering analysis and adjustment within the Chinese financial announcement document repository, we have identified a comprehensive array of tables. These include financial statements (such as balance sheets, income statements, cash flow statements, statements of shareholders' equity, and supplementary cash flow statements), notes to financial statements (covering accounts receivable, monetary funds, fair value disclosures, and payroll payable), as well as tables detailing directors, supervisors, senior management members, employee statistics, company profiles (encompassing basic information, contact details, principal accounting data, and financial ratios), shareholder information, and glossaries, among others.

We can extract style-related attributes from each table in the repository. These attributes include font information for each text line, line spacing, and alignment within text blocks contained in each cell. Additionally, we consider the alignment of text blocks within cells across rows or columns, the padding between aligned text blocks and their cell bounding boxes, the cells' background color, and the display mode, style, and color of the borders, both external and internal. For detailed illustrations, refer to the Appendix~\ref{appendix-Table-Components} and ~\ref{appendix-Table-Attributes}. These style-related attributes will be stored in the table's style profile, with specific examples provided in the Appendix~\ref{appendix-Table-style-profile}.

\subsection{Transformation of Table Style}

For certain datasets, such as FinTabNet, which already contain table structures and content data in HTML format, we can directly use this HTML data as source material to synthesize new tables through style transformation. In the case of other datasets, like those found in Chinese financial announcement document repositories, although the tables lack directly accessible structure and content information, many are bordered. This characteristic makes it relatively straightforward to deduce the tables' structure by identifying the lines within them and recognizing the text content in each cell. Consequently, these tables can likewise serve as source data for synthesizing tables with new styles.

Based on the content of the current source table, it is possible to match it to a corresponding category. This matching process can be achieved either by training a conventional text classifier or through the application of specific rules, depending on the context. Subsequently, one or more profiles are selected from the Style Profile Candidate Set of the matched category. Minor random adjustments are then made to certain attribute values within the selected profiles, which are ultimately defined as the style profiles for the target tables.

\subsection{Synthesis and Rendering of Target Tables}

Based on the style profile of the target table, coordinates are calculated for regular non-merged cells using a bottom-up approach (text line → text block → aligned text block bounding box → cell). For illustrations of table elements at various granularities and their corresponding attributes, see Appendices~\ref{appendix-Table-Components} and ~\ref{appendix-Table-Attributes}.
Based on the text line attributes defined in the style profile, calculate the height and width of all text lines. If a cell contains multiple text lines, the total height of the text block can be calculated based on the defined line spacing, along with the vertical relative coordinates of the top-left corner of each text line within the text block. Then, by considering the horizontal alignment of the text lines, calculate the total width of the text block and the horizontal relative coordinates of the top-left corner of each text line within the entire text block.
Scan each row of the table to identify the tallest text block among all regular non-merged cells and cells merged only in the horizontal direction in the current row. Use this height as the height for the aligned text block bounding box of that row. Calculate the vertical relative coordinates of the text block's top-left corner within the aligned text block bounding box, based on the text block's vertical alignment settings. Then, based on the cell's padding-top and padding-bottom attributes, calculate both the cell's height and the vertical relative coordinates of the aligned text block bounding box within the cell.
Similarly, scan each column of the table to identify the widest text block among all regular non-merged cells and cells merged only in the vertical direction in the current column. Use this width as the width for the aligned text block bounding box of that column. Calculate the horizontal relative coordinates of the top-left corner of the text block within the aligned text block bounding box, according to the text block's horizontal alignment settings. Then, using the cell's padding-left and padding-right attributes, determine the cell's width and the horizontal relative positioning of the aligned text block bounding box within the cell.

However, for merged cells or spanning cells, the calculation process is reversed, proceeding from top to bottom (cell → aligned text block bounding box → text block → text line) to compute coordinates.
Once the heights of all regular non-merged cells are determined, the height of vertically merged cells can also be obtained. Subsequently, based on the padding-top and padding-bottom attributes, the height of the aligned text block bounding box within this vertically merged cell can be determined. This allows for the calculation of the vertical relative coordinates of the aligned text block bounding box contained in the cell, with the vertical relative coordinates of the text block's top-left corner being calculated based on the text block's vertical alignment settings.
Similarly, once the widths of all regular non-merged cells are established, the width of horizontally merged cells can be obtained. Based on the padding-left and padding-right attributes, the width of the aligned text block bounding box within this horizontally merged cell is determined, enabling the calculation of the horizontal relative coordinates of the aligned text block bounding box contained in the cell. The horizontal relative coordinates of the text block's top-left corner within the aligned text block bounding box are calculated according to the text block's horizontal alignment settings.

Based on the cell sizes calculated in previous steps, the overall size of the table formed by stacking all cells can be determined. By connecting the border lines of cells in the same row and column, the absolute coordinates of the row and column dividers, i.e., the collection of horizontal and vertical lines of the table, are obtained. Calculating the absolute coordinates of the stacked cells, and then based on the relative coordinates of the aligned text block bounding box within the corresponding cell, the relative coordinates of the text block within the aligned text block bounding box, and the relative coordinates of the text lines within the corresponding text block, the absolute coordinates of the text lines within the table can ultimately be calculated.

Based on the table size calculated previously, create a blank canvas of the same dimensions. Then, using each cell's coordinates and background color, draw the table's background color. Next, render the corresponding text lines on the image canvas based on the coordinates, color, and font of each text line. Finally, draw the border lines using drawing tools based on the coordinates, mode, line type, and color of the borders. At the same time, output the annotation data in formats required by various table recognition models.

\section{Experiments}

To verify the practicality and effectiveness of the method proposed in this paper, we conducted experiments with tables from the financial announcement domain. This choice was made because financial announcements contain a large number of complex tables with varying styles. Recognizing these tables not only presents a significant technical challenge but also holds substantial real-world significance for various financial analysis applications.

Financial announcements in the United States have corresponding HTML format documents for their PDF files, and there already exists a large-scale English table annotation dataset like FinTabNet generated through automatic matching. However, Chinese financial announcements in PDF format do not have corresponding structured documents like HTML, and currently, there is no large-scale table annotation dataset available for them. In response to this situation, we have utilized the method proposed in this paper, leveraging the actual structure and content of tables within Chinese financial announcements, to generate the first large-scale table annotation dataset in the domain.

We collected a total of 5049 annual reports from Chinese listed companies in 2022, from which nearly 1.5 million tables were detected and extracted. The majority of these tables are bordered, with a minority being borderless. For comparison purposes, we sampled approximately the same magnitude of tables from these nearly 1.5 million tables as the English financial announcement dataset FinTabNet, totaling 105,600 bordered tables, to serve as the data source for table synthesis. To better recognize more complex tables with a greater number of merged cells, we increased the proportion of challenging complex tables with multiple merged or spanning cells during the sampling process, as shown in Table~\ref{table_sampled_distribution}.

\begin{table}
\centering
\caption{The data distribution of the 105,600 bordered tables sampled as the data source.}\label{table_sampled_distribution}
\begin{tabular}{|l|r|r|}
\hline
Spanning cell statistics &  Number & Percentage\\
\hline
no spanning cell &  52208 & 49.44\% \\
1 spanning cell &  4272 & 4.04\% \\
2 spanning cells & 13973 & 13.23\% \\
3 spanning cells & 15582 & 14.76\% \\
4 spanning cells or more & 19565 & 18.53\%\\
\hline
\end{tabular}
\end{table}

After conducting clustering analysis and manual inspection of the content of these 105,600 bordered tables, they were categorized into 14 categories. Our examination revealed that these source tables already encompass a very broad range of styles found in bordered tables within financial announcements. Therefore, when using these 105,600 bordered tables as the source for synthesizing new tables, 50\% of the tables were directly retained as part of the final synthesized table collection as bordered tables, without converting them into different styles of bordered tables.
Annotating borderless tables is more challenging than annotating bordered tables. In the synthesized dataset, there is a greater need to enhance support for recognizing borderless tables. Therefore, we transformed the other 50\% of the sampled 105,600 bordered tables into borderless tables.
%
%
The bordered tables serving as source tables can be relatively easily processed by identifying their border lines to extract the table's structure and content. Based on this content, the corresponding category is identified, and then a borderless style profile is selected from the Style Profile Candidate Set of that category. Subsequently, random adjustments of up to 10\% are made to certain style attribute values to generate the Style Profile for the synthesized target table. Finally, the target table image is synthesized and rendered, and annotation data is generated.

With the synthesized table annotation dataset available, we can train models for recognizing tables in financial announcements based on it. Here, we reproduced two recent deep learning-based end-to-end models, TableMaster~\cite{tablemaster-2021} and TableFormer~\cite{TableFormer-CVPR-2022}, both of which are based on an encoder-decoder architecture, particularly utilizing transformer-based decoders, hence both exhibit strong table recognition performance.
Compared to the implementations described in the original papers of TableMaster and TableFormer, we increased the maximum dimension of the input images to 640 pixels to accommodate the recognition of more complex tables.
%
TableMaster and TableFormer have very similar underlying architectures. In TableFormer, the Transformer has fewer layers and heads, and it also employs adaptive pooling to reduce the size of the feature map output by the CNN Backbone. Consequently, TableMaster has a higher number of parameters and, correspondingly, a higher computational complexity. Our experiments also show that TableMaster's overall performance is better than that of TableFormer.
Given that the experimental results and conclusions of both models are consistent, and in the interest of conserving space and presenting the information more concisely, we only report the experimental results of TableMaster in the subsequent sections.

We employ the Tree-Edit-Distance-based Similarity (TEDS)~\cite{PubTabNet-ECCV-2020}, a metric commonly used in table structure recognition literature, to evaluate the performance of table structure recognition. TEDS assesses the similarity between the tree structures of tables. To utilize the TEDS metric, tables must be represented as tree structures in HTML format.
Considering that accounting for errors in the text content of tables could result in unfair comparisons due to the varied text extraction methods or OCR models employed by different table recognition methods, we utilize a modified version of TEDS, named TEDS-Struct. This version focuses on the accuracy of table structure recognition, explicitly disregarding the outcomes from text extraction or OCR processes. 
We also investigate the performance of text block detection (AP50, MS COCO AP at IoU=.50)~\cite{VAST-CVPR-2023}, which is crucial for the precise matching of each cell to its corresponding text content.


We sampled 2,290 real tables from Chinese financial announcements, ensuring no overlap with the 105,600 tables previously sampled as sources for synthesizing table structures and content. Among these, 1,000 are bordered tables, and 1,290 are borderless tables, aiming to increase focus on borderless tables. Additionally, the selection of tables also prioritized more challenging tables that include multiple merged or spanning cells; for specifics, please refer to Table~\ref{table_benchmark_distribution}. Using table recognition models trained on synthesized data, we conducted recognition and automatic annotation on these 2,290 tables, followed by manual verification to create the first benchmark for complex tables in the Chinese financial announcement domain. This benchmark can be used to evaluate the table recognition performance of models trained on synthesized data, thereby demonstrating the quality of the synthesized data.

\begin{table}
\centering
\caption{The data distribution of real-world table benchmark in the Chinese financial announcement domain.}\label{table_benchmark_distribution}
\begin{tabular}{|l|r|r|r|r|r|r|}
\hline
\multirow{2}{*}{Spanning cell statistics} & \multicolumn{2}{c|}{All} & \multicolumn{2}{c|}{Bordered} & \multicolumn{2}{c|}{Borderless} \\ \cline{2-7} 
 & Number & Percentage & Number & Percentage & Number & Percentage \\ 
\hline
No spanning cell & 1045 & 45.63\% & 442 & 44.20\% & 603 & 46.74\% \\ 
1 spanning cell & 82 & 3.58\% & 59 & 5.90\% & 23 & 1.78\% \\ 
2 spanning cells & 268 & 11.70\% & 145 & 14.50\% & 123 & 9.53\% \\ 
3 spanning cells & 373 & 16.29\% & 100 & 10.00\% & 273 & 21.16\% \\ 
4 spanning cells or more & 522 & 22.79\% & 254 & 25.40\% & 268 & 20.78\% \\ 
\hline
\end{tabular}
\end{table}

The FinTabNet table dataset, extracted from English financial announcements, was annotated through automatic matching between PDF and HTML, resulting in a considerable number of errors.
The FinTabNet training set includes a large number of tables, and due to time constraints, corrections were not conducted. However, for the FinTabNet test set, which contains 10,635 tables, we manually reviewed and used automated scripts to correct 3,733 tables with inconsistent annotations of leader dots. Additionally, we removed 954 tables that had errors in table positioning, structural annotations, or in text box annotations or shifts. The presence of these errors, resulting from automatic annotation, does not imply that the tables are structurally more complex; thus, removing these tables did not decrease the overall difficulty of the FinTabNet table recognition evaluation task. After our review, the corrected FinTabNet test set now comprises 9,681 tables.

Table~\ref{table_results_benchmark} presents the evaluation results of TableMaster, trained on the synthesized dataset, on our Chinese financial announcement table benchmark dataset. In comparison to the results listed in Table~\ref{table_result_corrected_FinTabNet}—where TableMaster were trained on the FinTabNet training set and evaluated on the corrected FinTabNet test set—the performance is relatively lower. This discrepancy is attributed to the tables extracted from Chinese financial announcements generally being more complex than those from the FinTabNet dataset, which extracts tables from English financial announcements. For instance, tables from Chinese announcements often contain more cells with multi-line text blocks, cells with very long texts, or a higher density of cells. These tables frequently feature structurally complex spanning cells, significant horizontal alignment deviations of text blocks within the same column, among other factors, all of which significantly increase the difficulty of table recognition in Chinese financial announcements. Furthermore, there is still room for improvement in predicting the positions of text block bounding boxes, which could further enhance the accuracy in matching the recognized table structure with the text content~\cite{VAST-CVPR-2023}.

\begin{table}
\centering
\caption{The evaluation results for TableMaster, after being trained on the synthesized dataset, on the Chinese financial announcement table benchmark dataset.}\label{table_results_benchmark}
\begin{tabular}{|l|r|r|r|}
\hline
Spanning Cell Statistics &  TEDS & TEDS-Struct & AP-50 \\
\hline
 
 all tables & 0.9091  &  0.9579 & 0.482 \\  
 \hline
 no spanning cell & 0.9216  & 0.9632 & 0.501\\
 1 spanning cell & 0.9131  & 0.9503 & 0.568\\
 2 spanning cells & 0.9375 & 0.9696 & 0.592 \\
 3 spanning cells & 0.8785 & 0.9449 & 0.505\\
 4 spanning cells or more & 0.8908 & 0.9515 & 0.458\\ 
  \hline
 merged on rows \& columns & 0.8906 & 0.9500 & 0.497\\
\hline
\end{tabular}
\end{table}


\begin{table}
\centering
\caption{The evaluation results for TableMaster, after being trained on the FinTabNet training set, on the corrected FinTabNet test set.}\label{table_result_corrected_FinTabNet}
\begin{tabular}{|l|r|r|r|}
\hline
 Spanning cell statistics &  TEDS & TEDS-Struct & AP-50 \\
\hline
all tables & 0.9758  &  0.9856 & 0.619 \\ 
\hline
no spanning cell & 0.9727  & 0.9829 & 0.631\\
1 spanning cell & 0.9830  & 0.9922 & 0.654\\
2 spanning cells & 0.9818 & 0.9896 & 0.630 \\
3 spanning cells & 0.9657 & 0.9783 & 0.557\\
4 spanning cells or more & 0.9342 & 0.9552 & 0.511\\  
 \hline
merged on rows \& columns & 0.9503 & 0.9620 & 0.574\\
\hline
\end{tabular}
\end{table}

To further validate the practicality of our method, we applied the data synthesis method proposed in this paper to augment the training data for FinTabNet. 
The original FinTabNet training dataset contains relatively few tables with multiple spanning cells. In augmenting the dataset while keeping the total number of tables roughly the same, a portion of the augmented dataset was obtained directly by sampling from the original FinTabNet training dataset, and another portion was synthesized using the method described in this paper. The augmented dataset increases the proportion of tables containing multiple spanning cells, with specific distribution information of the tables available in Table~\ref{table_augmented_distribution}.
Table~\ref{table_result_augmented_corrected_FinTabNet} presents the experimental results on the corrected FinTabNet test dataset of models trained using the augmented FinTabNet training data. Compared with Table~\ref{table_result_corrected_FinTabNet}, it shows that data augmentation using the method proposed in this paper can comprehensively improve the performance of FinTabNet table recognition, especially in recognizing more complex tables with multiple spanning cells. This demonstrates the effectiveness of our table synthesis method for practical application scenarios.

\begin{table}
\centering
\caption{The data distribution of Augmented FinTabNet training dataset.}\label{table_augmented_distribution}
\begin{tabular}{|l|r|r|r|r|r|r|}
\hline
\multirow{2}{*}{Table Type} &  \multirow{2}{*}{FinTabNet} & \multicolumn{3}{c|}{Augmented dataset} \\ \cline{3-5}
& & Sampled & Synthesized & Sum \\
\hline
no spanning cell & 44k+  & 10000 & 10000 & 20000\\
1 spanning cell & 22k+  & 7500 & 7500 & 15000\\
2 spanning cells & 14k+  & 14439 & 561 & 15000\\
3 spanning cells &  4k+ & 4824 & 10176 & 15000\\
4 spanning cells or more &  5k+ & 5026 & 14711 & 19737\\
 \hline
merged on rows \& columns & 2k+ & 263  & 7954 & 8217 \\
\hline
\end{tabular}
\end{table}


\begin{table}
\centering
\caption{The evaluation results for TableMaster, after being trained on the Augmented FinTabNet training set, on the corrected FinTabNet test set.}\label{table_result_augmented_corrected_FinTabNet}
\begin{tabular}{|l|r|r|r|}
\hline
Spanning cell statistics &  TEDS & TEDS-Struct & AP-50 \\
\hline 
all tables & 0.9847  &  0.9971 & 0.789 \\ 
\hline
 no spanning cell & 0.9805  & 0.9971 & 0.746\\
 1 spanning cell & 0.9895  & 0.9982 & 0.813\\
 2 spanning cells & 0.9912 & 0.9979 & 0.827 \\
 3 spanning cells & 0.9794 & 0.9906 & 0.809\\
 4 spanning cells or more & 0.9740 & 0.9906 & 0.747\\  
  \hline
 merged on rows \& columns & 0.9684 & 0.9794 & 0.841\\
\hline
\end{tabular}
\end{table}

\section{Conclusion and Future Work}

Unlike previous methods that rely on automatic matching between PDFs and structured source codes or on random synthesis of table structures and content, this paper introduces a novel approach for synthesizing high-quality tables and annotated data. This method leverages the structure and content of existing real tables to replicate authentic table styles of the target domain. We applied this approach within the financial sector, producing the first extensive table annotation dataset for the Chinese financial announcement domain and enhancing the English financial table dataset, FinTabNet. Our experiments demonstrate the real-world applicability and effectiveness of this table synthesis method.

The experiments utilized end-to-end table recognition methods such as TableMaster, chosen for their relatively simple data preparation process and proven excellence in performance across previous studies. However, these methods encounter considerable challenges when recognizing the structurally complex tables common in Chinese financial announcements. Future work will explore additional methods, particularly those based on segmentation and merging~\cite{SPLERGE-2019}, which are anticipated to yield improved results for tables with intricate spanning cell structures. Employing a diverse range of table recognition methods will enable a more thorough assessment of the synthesized data's quality.

Furthermore, we aim to broaden the diversity of table styles by incorporating tables from audit reports, to publicly release our extensive synthesized dataset for the Chinese financial announcement domain, and to enlarge and publicly unveil the benchmark for real complex tables within the same domain.
%
%
%
 \bibliographystyle{splncs04}
 \bibliography{ref}
%






\section{Appendix}

\subsection{Table Elements at various granularities}\label{appendix-Table-Components}

\subsubsection{Cells}
\begin{figure}
\includegraphics[width=0.8\textwidth]{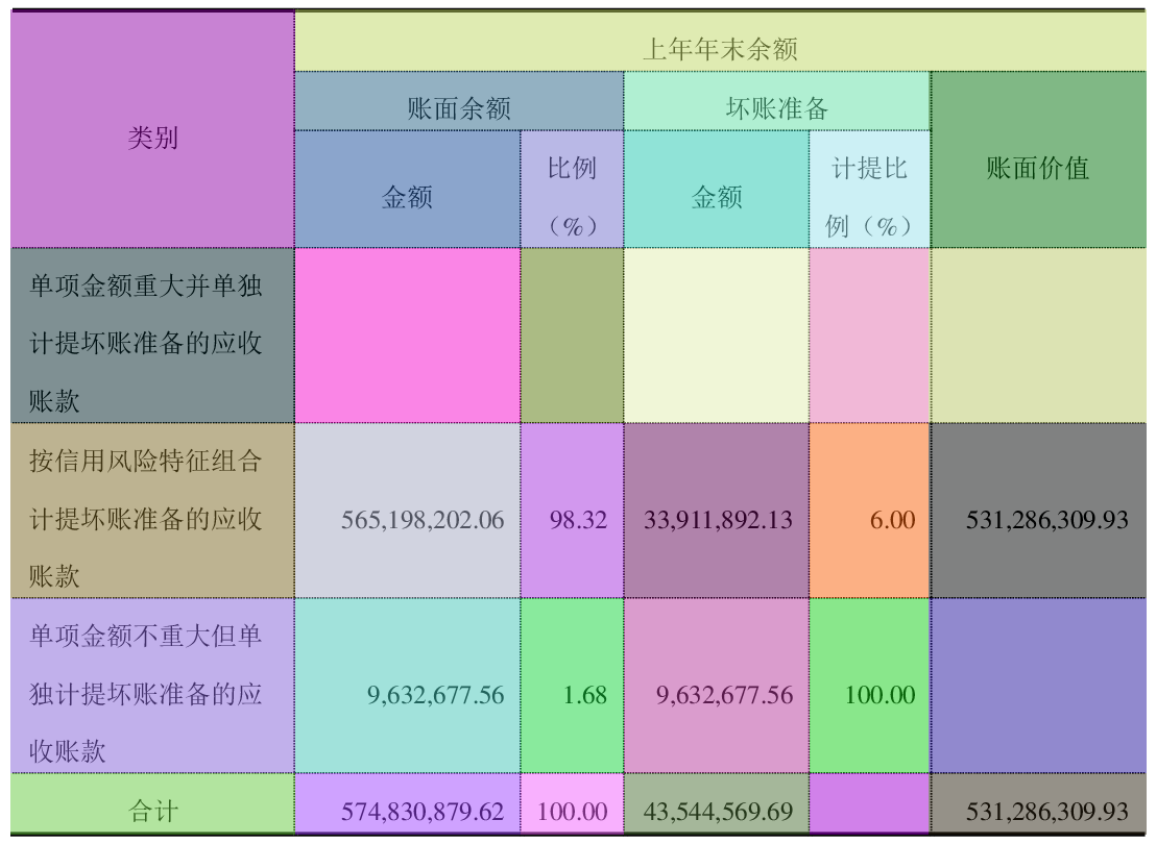}
\caption{Illustration of table cells.} \label{fig-cells}
\end{figure}

In Fig~\ref{fig-cells}, each area filled with a different color represents a cell. Only tables with complete border lines have definite cell coordinates.

\subsubsection{Text Lines}
\begin{figure}
\includegraphics[width=0.8\textwidth]{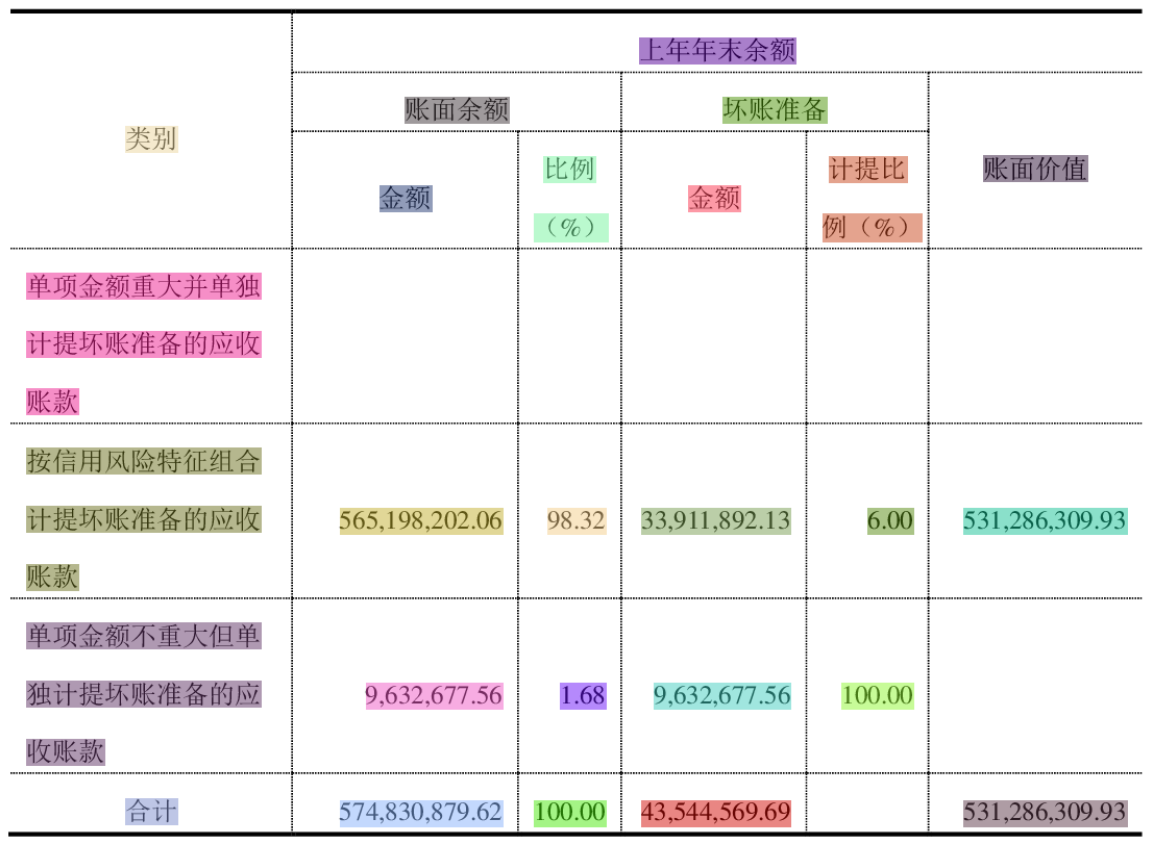}
\caption{Illustration of text lines.} \label{fig-text-lines}
\end{figure}

In Fig.~\ref{fig-text-lines}, the areas filled with color represent text lines. Text lines within the same cell share the same color, and a single cell often contains multiple text lines.

\subsubsection{Text Blocks}
\begin{figure}
\includegraphics[width=0.8\textwidth]{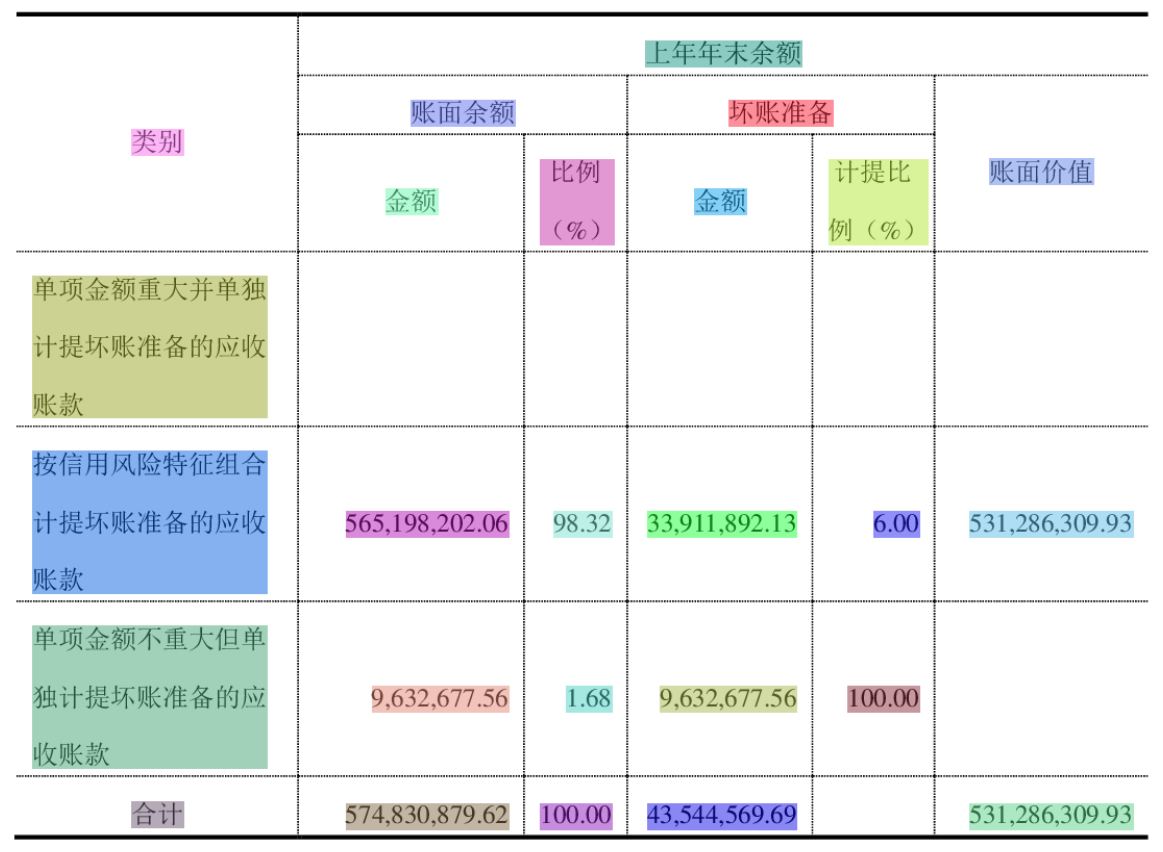}
\caption{Illustration of text blocks.} \label{fig-text-blocks}
\end{figure}

In Fig.~\ref{fig-text-blocks}, each area filled with a different color represents a text block, and the bounding box of a text block covers all text lines within the same cell.

\subsubsection{Aligned Text Block Bounding Boxes}
\begin{figure}
\includegraphics[width=0.8\textwidth]{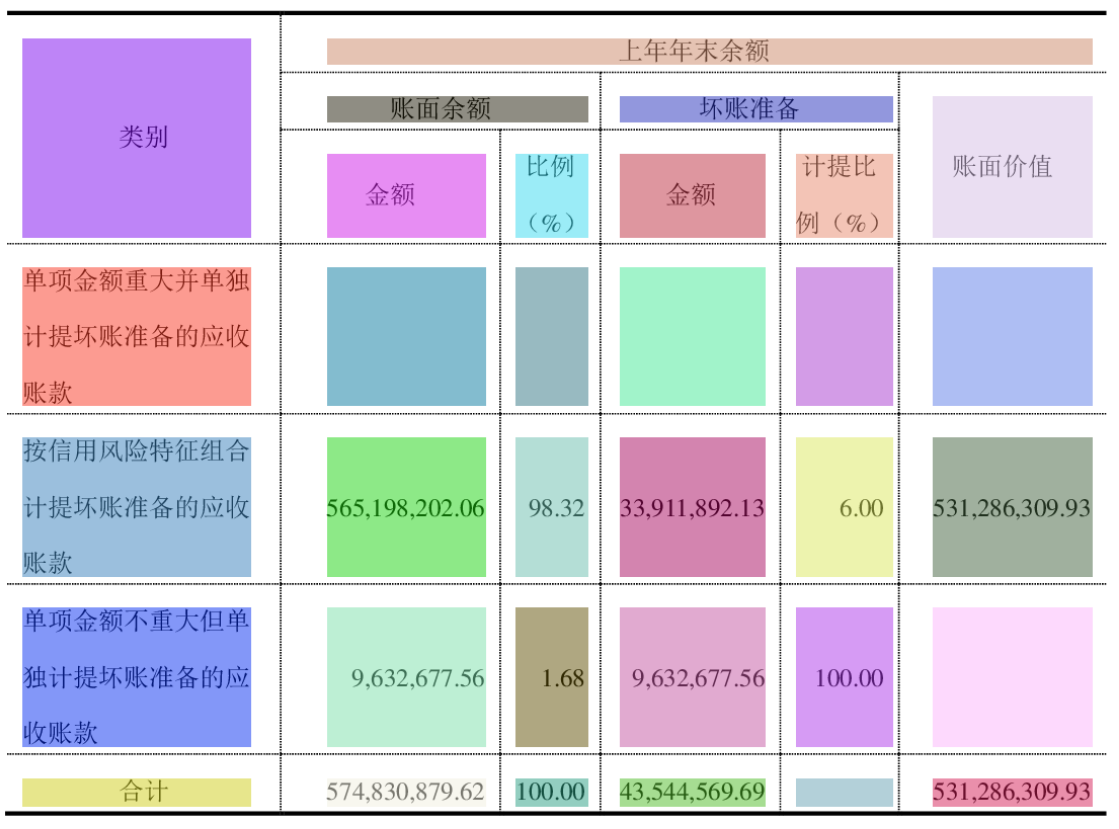}
\caption{Illustration of Aligned Text Block Bounding Boxes.} \label{fig-aligned-bounding-box-text-blocks}
\end{figure}

In Fig.~\ref{fig-aligned-bounding-box-text-blocks}, each area filled with a different color is an aligned text block bounding box.
Identify the highest upper bound and the lowest lower bound from all text blocks among all regular non-merged cells and cells merged only in the horizontal direction in the current row to calculate the height, and use this height as the height of the aligned text block bounding box for that row.
Identify the leftmost boundary and the rightmost boundary from all text blocks among all regular non-merged cells and cells merged only in the vertical direction in the current column to calculate the width, and use this width as the width of the aligned text block bounding box for that column.
The width of the aligned text block bounding box for a horizontally merged cell is determined by the left boundary of the aligned text block bounding box of the leftmost column involved in the merge and the right boundary of the aligned text block bounding box of the rightmost column involved in the merge.
The height of the aligned text block bounding box for a vertically merged cell is determined by the upper boundary of the aligned text block bounding box of the topmost row involved in the merge and the lower boundary of the aligned text block bounding box of the bottommost row involved in the merge.

\subsection{Attributes of Table Elements} \label{appendix-Table-Attributes}

\subsubsection{Attributes of Text Lines and Text Blocks}

\begin{figure}
\includegraphics[width=0.8\textwidth]{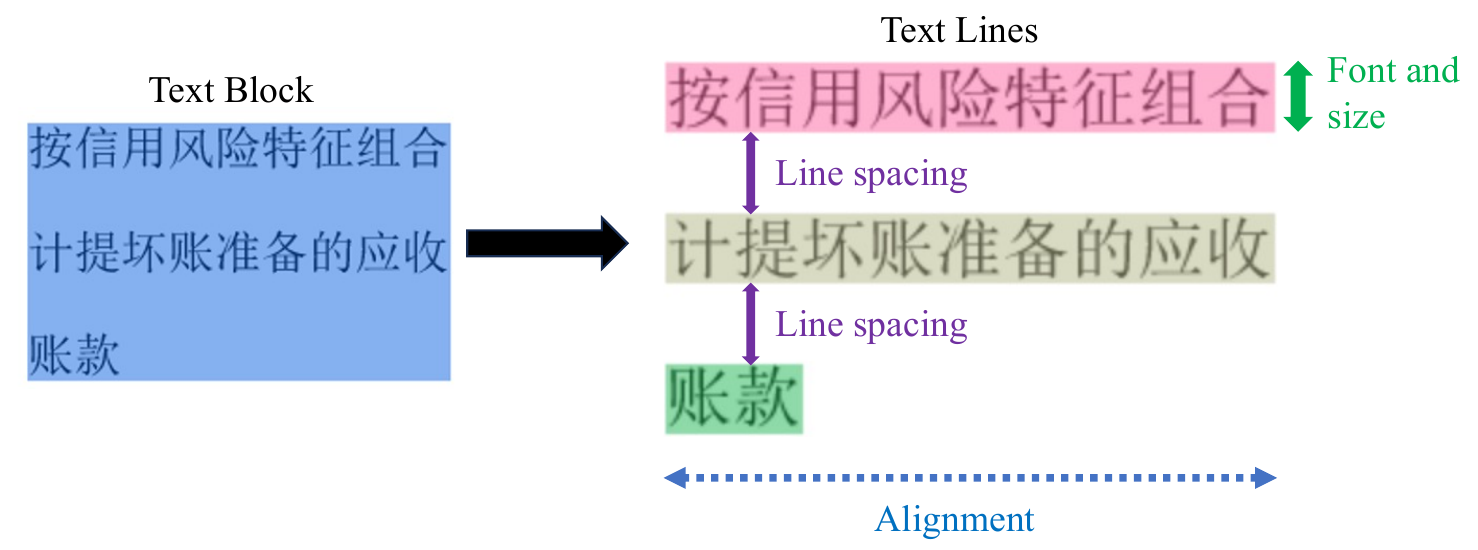}
\caption{Illustration of Attributes of text lines and text blocks.} \label{fig-attributes-text-lines-blocks}
\end{figure}

The attributes of text lines include the font, size, and color of the text, all of which can be set at various granularities such as table, row, column, or cell.
The attributes of text blocks include the line spacing between the contained text lines and the alignment of text lines within the text block (centered/left-aligned/right-aligned/specified indent distance), all of which can be set at various granularities such as table, row, column, or cell.
Fig.~\ref{fig-attributes-text-lines-blocks} shows an illustration of the attributes of a text block and the text lines contained within it, extracted from Fig.~\ref{fig-text-blocks}.
%

\subsubsection{Attributes of Cells and corresponding Aligned Text Block Bounding Boxes}
\begin{figure}
\includegraphics[width=0.8\textwidth]{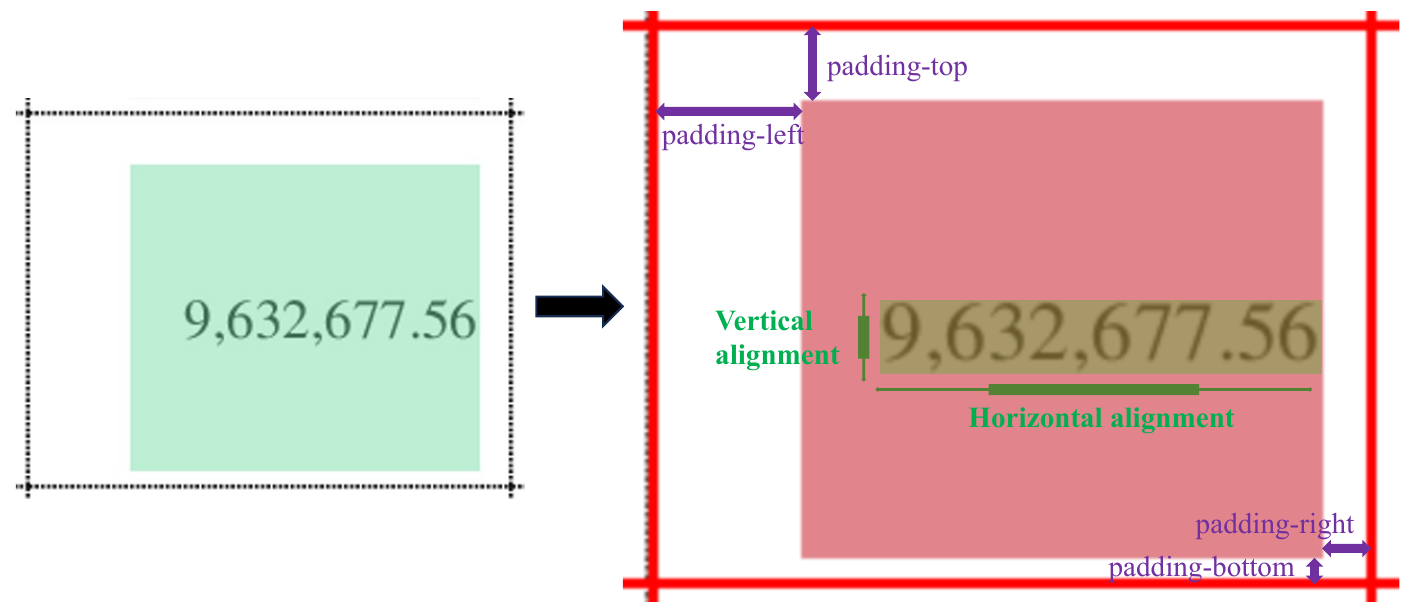}
\caption{Illustration of Attributes of Cells and corresponding Aligned Text Block Bounding Boxes.} \label{fig-attribute-cells}
\end{figure}

The attributes of an aligned text block bounding box mainly involve the alignment of the contained text block, including horizontal alignment (left-aligned, right-aligned, centered, indented by a specified distance to the left or right) and vertical alignment (top-aligned, bottom-aligned, centered, indented by a specified distance from the top or bottom). All of these can be set at various granularities such as table, row, column, or cell.
The attributes of a cell include the padding, defined as the distance between the cell bounding box and the aligned text block bounding box, as well as the cell's background color. All of these can be set at various granularities such as table, row, column, or cell.
Fig.~\ref{fig-attribute-cells} displays a cell and its corresponding aligned text block bounding box, as extracted from Fig.~\ref{fig-aligned-bounding-box-text-blocks}. 
%

\subsubsection{Attributes of borders}

In tables, borders are categorized into outer and inner borders. The attributes of outer borders apply to the table as a whole, while inner border attributes can be adjusted at various levels of granularity, such as by table, row, column, or cell.
Outer border modes can be fully visible, absent on the sides, absent on the top and bottom, or completely absent. Inner border options encompass fully visible, absent horizontal lines, absent vertical lines, completely absent, partially absent horizontal lines, and partially absent vertical lines. Outer border line types include options like single solid lines and double solid lines, along with specifications for their thickness. Inner border line types feature solid lines, dashed lines, and details regarding line thickness and the spacing in dashed lines.
Furthermore, colors can be specified for both outer and inner borders.

\subsection{Table Style Profile} \label{appendix-Table-style-profile}

\begin{figure}
\includegraphics[width=0.8\textwidth]{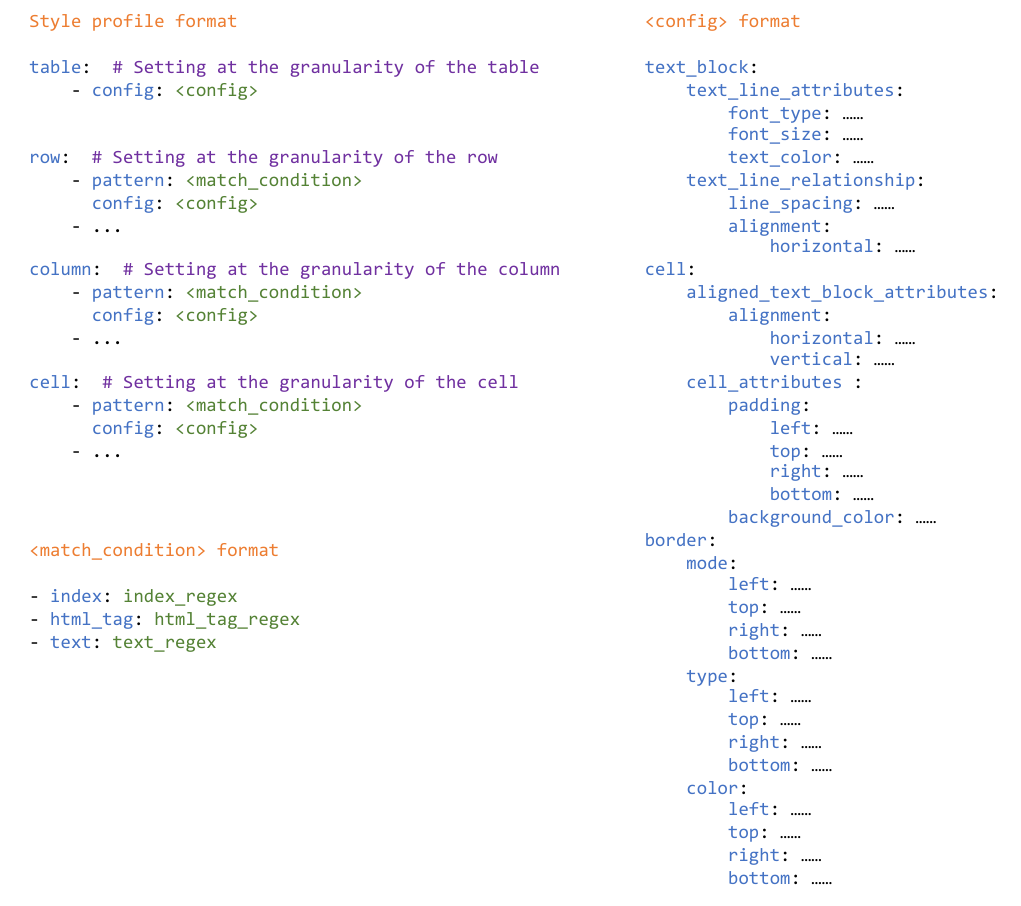}
\caption{Illustration of an example of the format of a Table Style Profile.} \label{figure-profile-template}
\end{figure}

Fig.~\ref{figure-profile-template} shows an example of the format of a Table Style Profile.

\end{document}